\documentclass{article}
\usepackage{spconf,amsmath}
\usepackage{cite}
\usepackage{hyperref}

\usepackage{graphicx}
\usepackage{newfloat}
\usepackage{xcolor}
\usepackage{siunitx}
\usepackage{collcell}
\usepackage{graphicx} 

\usepackage{tikz}
\usepackage{subcaption}
\usepackage{mathtools}
\usepackage{url}
\usepackage[font=small]{caption}
\usepackage[utf8]{inputenc}         
\usepackage{wrapfig}

\usepackage[full]{complexity}

\usepackage{polski}
\usepackage{xcolor}

\usepackage[T1]{fontenc}  
\usepackage{amsmath,amsfonts,amssymb,amsthm}

\title{Optimizing Kernel-Target Alignment for cloud detection in multispectral satellite images}

\name{\begin{tabular}{c}Artur Miroszewski$^{1}$, Jakub Mielczarek$^1$, Filip Szczepanek$^1$, Grzegorz Czelusta$^1$,\\{Bartosz Grabowski}$^{2}$, \textit{Bertrand Le Saux}$^3$, and \textit{Jakub Nalepa}$^{2,4}$\end{tabular}}

\address{$^1$Jagiellonian University, prof. S. \L{}ojasiewicza 11, 30-348 Cracow, Poland\\$^2$KP Labs, Konarskiego 18C, 44-100 Gliwice, Poland\\$^3$European Space Agency, Largo Galileo Galilei 1, 00044 Frascati, Italy\\$^4$Silesian University of Technology, Akademicka 16, 44-100 Gliwice, Poland
\\ \texttt{artur.miroszewski@uj.edu.pl, jnalepa@ieee.org}}

\begin{document}
%
\maketitle
\begin{abstract}
The optimization of Kernel-Target Alignment (TA) has been recently proposed as a way to reduce the number of hardware resources in quantum classifiers. It allows to exchange highly expressive and costly circuits to moderate size, task oriented ones. In this work we propose a simple toy model to study the optimization landscape of the Kernel-Target Alignment. We find that for underparameterized circuits the optimization landscape possess either many local extrema or becomes flat with narrow global extremum.
We find the dependence of the width of the global extremum peak on the amount of data introduced to the model. The experimental study was performed using multispectral satellite data, and we targeted the cloud detection task, being one of the most fundamental and important image analysis tasks in remote sensing.
\end{abstract}

\section{Introduction}

The highly expressive quantum and hybrid classical-quantum machine learning algorithms are known for being notoriously hard to train. 
The simulation of optimization in quantum Hilbert space is a computationally intensive task, frequently demanding access to high-performance computing resources. Additionally, we encounter problems in which the scaling of computational complexity technically wastes the possibility of obtaining satisfactory results.
Quantum neural networks (QNNs) are known to suffer from Barren Plateau phenomenon \cite{mcclean2018barren}, while kernel-based methods, like hybrid support vector machines (SVMs)~\cite{9554802,electronics10202482}, experience value concentration in kernel entries \cite{thanasilp2022exponential}.
To avoid the aforementioned problems, researchers tend not to over-parameterize when designing quantum circuits. 
This can be done either by focusing solely on the highly symmetric problems or by introducing additional hyperparameters of the circuits, which are then adjusted according to some measure of quality of the resulting quantum map.
As real-life data rarely does possess explicit symmetries, we are usually compelled to optimize circuit designs.
In the kernel-based classification algorithms, the measure to assess the quality of the quantum map is the Kernel-Target Alignment (TA) \cite{cristianini2001kernel, hubregtsen2022training}.
In this paper, we propose a simple, exactly solvable toy-model for the TA landscape examination for models of low expressivity. We show that the TA landscape becomes harder to optimize not only with the growing complexity of the structure in the training data, but also with the amount of training data itself. The landscape becomes flat, the width of global extremum and the TA expected value decrease.
Finally, we confront the toy-model's predictions with the scenario based on real-life multispectral satellite data. Here, we exploit the Landsat-8 38-Cloud dataset~\cite{38-cloud-1, 38-cloud-2} of multispectral images, and target the binary problem of cloud detection in such imagery. This task is the ``hello, world'' in satellite image data analysis, and may be considered a smart data compression and selection step in the processing chain to prune cloudy, hence useless images from further processing. Our experimental study showed that, for the underparameterized circuits, the TA optimization becomes harder as we uncover the underlying structure of the available training data. 

\section{Toy model}

\begin{figure*}[ht]
    \centering
    \includegraphics[height=5 cm]{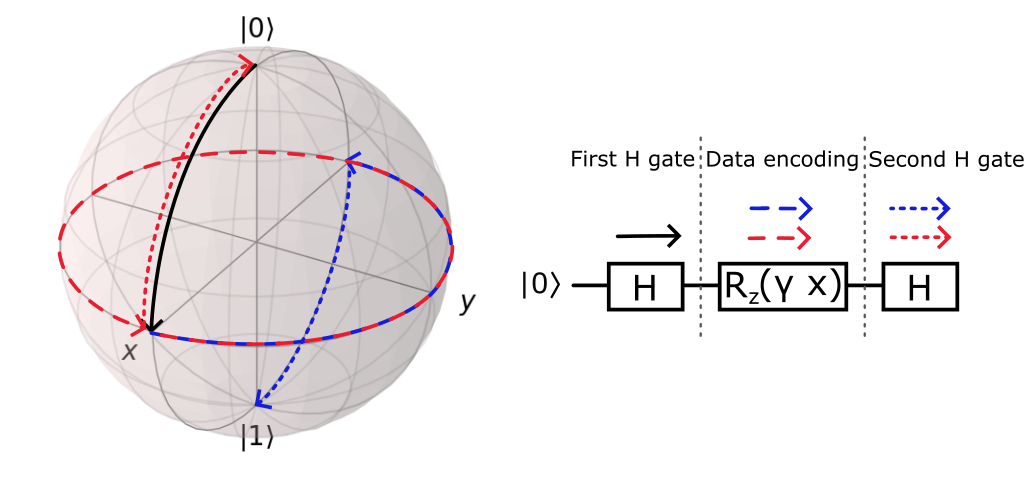}
    \caption{The feature map proposed for a toy model. First, it transforms the initial $|0\rangle$ to the $|+\rangle$ state which resides on the Bloch spheres' equator. Then the parameterized rotation is applied is order to move the state around the equator. The last Hadamard gate transforms us back to the measurement basis. For the specific data set used in the toy model, the rotation with parameter $\gamma = \pi \cdot (N-1)$ sends the states of opposite classes to $|+\rangle$ and $|-\rangle$ states. After the final Hadamard, the points are encoded on north and south pole of the Bloch sphere.}
    \label{fig:feature_map}
\end{figure*}

We propose a toy model for the study of Kernel-Target alignment landscape. It consists of the two-class labeled dataset and a parameterized feature map. Although being simple, the toy model is not trivial and captures some of the important challenges of the TA optimization. Even though the model utilizes one qubit and one feature map parameter, it can be seen as a basic building block for all non-entangling, hence separable, $n$-qubit feature maps. Therefore the optimization issues found in this work can be found representative in a wide class of quantum, kernelized problems. On the other hand the simplicity of the model allows for obtaining analytical results and clear, visual inspection of its features.

Let us consider a synthetic set consisting of $N$ equidistant data points on the interval $[0,1]$, $x_i = \frac{i-1}{N-1},\ i=1,\dots, N$. The classes for the points are alternating $y_i = (-1)^{i-1}$. For $N$ even we always have a balanced dataset in the studied model.   
In order to encode the data on the Bloch sphere, we use a simple feature map consisting of a Hadamard gate followed by a parameterized Z-rotation $R_Z(\gamma x)$ and another Hadamard gate (Fig. \ref{fig:feature_map}). The circuit effectively moves the initial state to the spheres' equator, rotates the state by a parameterized angle and transforms it back to the computational basis. One can obtain a perfect separation of the data points, when $+1$ ($-1$) class occupies north (south) pole of the Bloch sphere for $\gamma = (N-1) \cdot \pi$. 

The feature map induces a fidelity kernel of the form:
\begin{equation}\label{eq:kernel_entry}
    \mathcal{K}_{ij} = |\langle x_j | x_i \rangle|^2 = \cos^2\left[\frac{\gamma}{2}(x_i - x_j) \right].
\end{equation}
The kernel entries can be estimated by running a parameterized circuit on a quantum computer or simulated on the classical computer.

\section{Kernel-Target Alignment landscape}
The TA, being the measure which quantifies the separation of data points belonging to different classes, is given as~\cite{cristianini2001kernel}:
\begin{equation}\label{eq:target_alignment}
\mathcal{T}(\mathcal{K}) = \frac{\langle \mathcal{K}, \bar{\mathcal{K}} \rangle_F}{\sqrt{\langle \mathcal{K}, \mathcal{K} \rangle_F \langle \bar{\mathcal{K}}, \bar{\mathcal{K}} \rangle_F}},
\end{equation}
where $\mathcal{K}$ is a kernel matrix, $\bar{\mathcal{K}}_{ij} = y_i y_j$ is an ideal kernel and  $\langle A, B\rangle_F = Tr\{ A^T B \}$ is a Frobenius inner product. 
We expect that with the greater value of TA, the classification accuracy increases, as data points belonging to different classes are well separated in the feature space.
As the kernel function (Equation~\ref{eq:kernel_entry}) is non-negative the maximum value of the Kernel-Target Alignment is $\mathcal{T}(\mathcal{K^*}) = \frac{1}{\sqrt{2}}$, for $K^*$ consisting of vanishing entries for points in different classes and unity values for points in the same class. In general, one can center the dataset \cite{scholkopf1998nonlinear}, to rescale the $\mathcal{T}(\mathcal{K})$ to the interval $[0,1]$. However this procedure would not impact the results of the work, therefore we omit this step.
Let us investigate the Kernel-Target Alignment function for the proposed toy model. Firstly, for a balanced dataset one can write an ideal kernel normalization $\langle \bar{\mathcal{K}},\bar{\mathcal{K}} \rangle_F = N^2$.
Knowing that the even (odd) indices of data points $x_i$, $i = 2l$ ($i = 2l-1$), $l \in \{1,\dots, \frac{N}{2}\}$ have $-1$ ($+1$) label, one can obtain the following form of Kernel-Target Alignment in the toy model:
\begin{equation}\label{eq:TKA_toy_model}
\mathcal{T}(\mathcal{K}) = \frac{1}{N}\frac{\langle \mathcal{K}, \bar{\mathcal{K}} \rangle_F}{\sqrt{\langle \mathcal{K}, \mathcal{K} \rangle_F}},
\end{equation}
where
\begin{equation}
    \langle \mathcal{K}, \bar{\mathcal{K}} \rangle_F = 2 \sum_{\alpha \in \{0,1\}} \sum_{k,l=1}^{N/2} (-)^{\alpha} \cos^2\left[ \frac{\gamma (k-l+\alpha/2)}{N-1} \right],
\end{equation}
and
\begin{equation}
    \langle \mathcal{K}, \mathcal{K} \rangle_F^2=2\sum_{\alpha \in \{0,1\} } \sum_{k,l=1}^{N/2} \cos^4\left[ \frac{\gamma(k-l+\alpha/2)}{N-1}\right].
\end{equation}

\begin{figure*}[ht]
        \centering
           \subfloat[]{%
              \includegraphics[height=4cm]{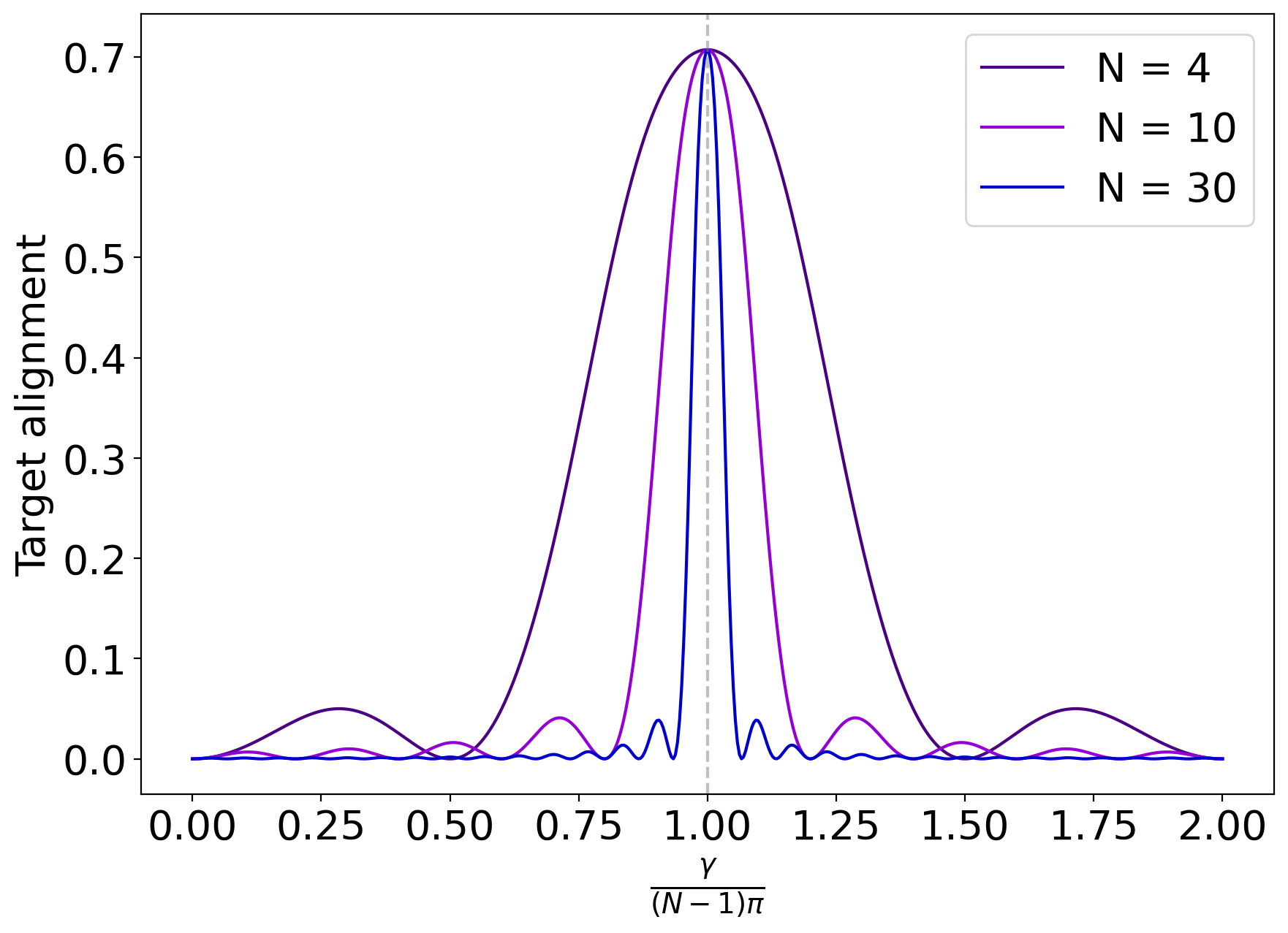}%
              \label{fig:TA_landscape}%
           } 
           \subfloat[]{%
              \includegraphics[height=4cm]{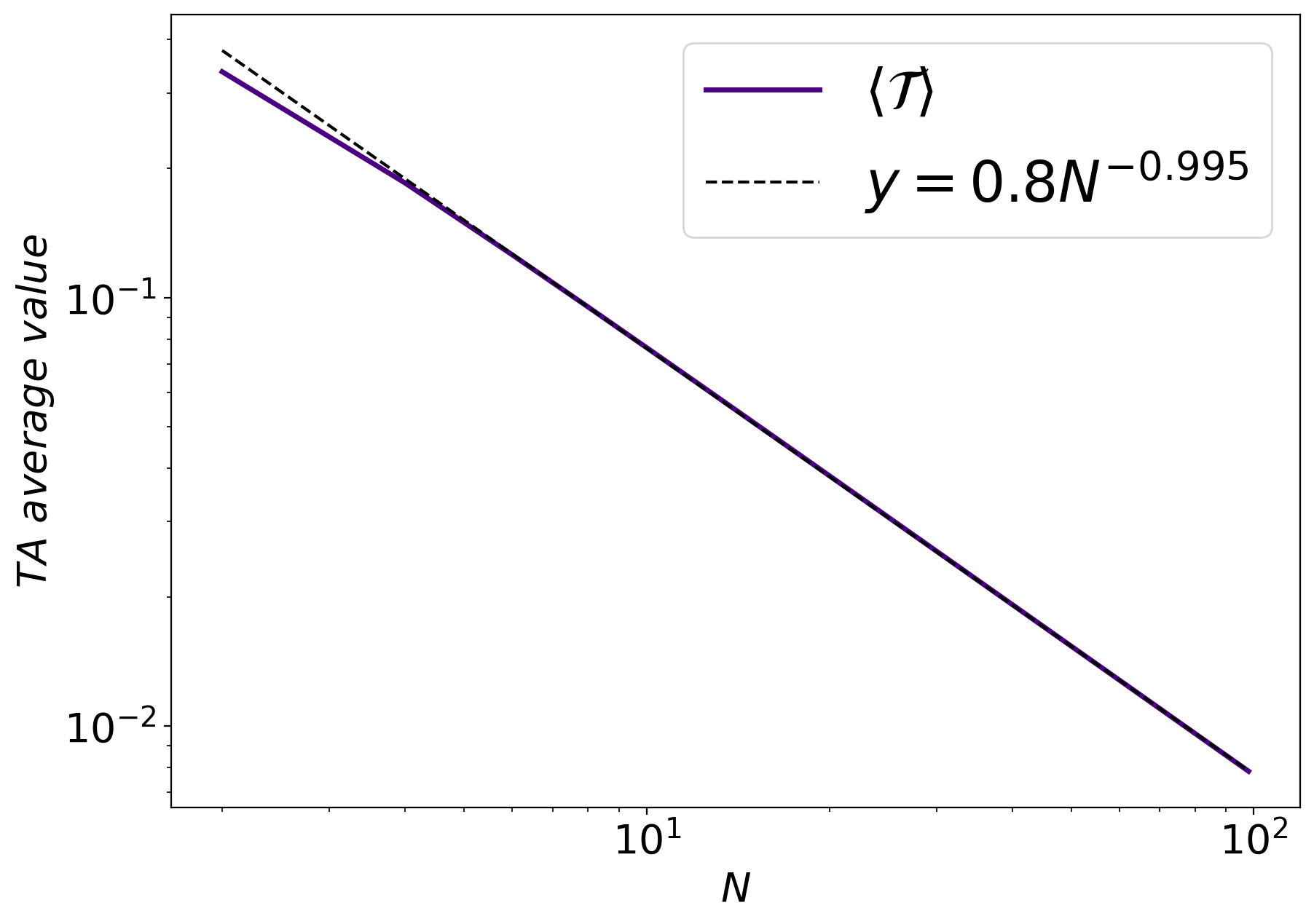}%
              \label{fig:fit}%
           }
           \subfloat[]{%
              \includegraphics[height=4cm]{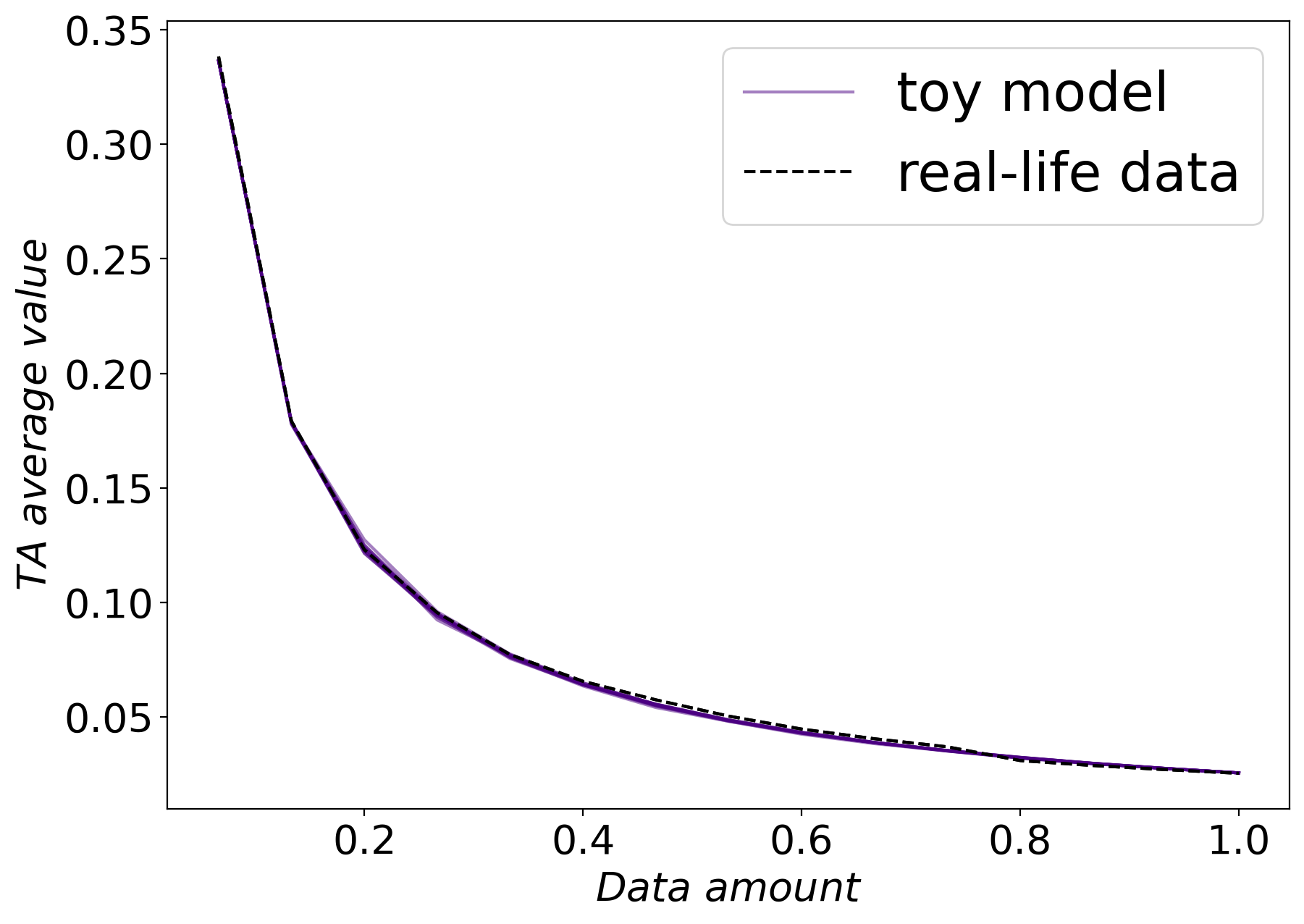}%
              \label{fig:real_av}%
           }
           \caption{Kernel-Target Alignment for the toy model and 38-Clouds data. a) An optimization landscape for the toy's model TA. The $\gamma$ parameter governs the Z-rotation in the quantum map. The optimal value is $\hat{\gamma}=(N-1) \cdot \pi$. 
           b) The average value of the Target-Kernel Alignment landscape as a function of a data set size $N$. The dashed line represents a fit confirming the scaling obtained in gaussian approximation (\ref{eq:gaussian}). 
           c) The expected value of TA as we introduce training data points in the toy model and points from the 38-Cloud dataset. For each model 10 simulations are presented on the plot.}
           \label{fig:default}
    \end{figure*}

From the above formulas one can infer that the $\mathcal{T}(\mathcal{K})$ function is periodic in $\gamma$ parameter with $2\pi\cdot(N-1)$ period. In Fig. \ref{fig:TA_landscape} we show one-period of Kernel-Target Alignment landscape for different values of $N$. The peak centered at solution solution $\gamma = \pi \cdot (N-1)$ is visible in the middle.

One can immediately see that the width of the central peak decreases with the increasing amount of points $N$. By expanding the function (Equation~\ref{eq:TKA_toy_model}) around the maximum to the second order and fitting a Gaussian to it, one obtains:
\begin{equation}\label{eq:gaussian}
\begin{split}
        &\mathcal{G}_{\gamma}(\mu,\sigma) = \frac{1}{\sqrt{2}}e^{-\frac{1}{2}\left(\frac{\gamma-\mu}{\sigma}\right)^2},\\ 
        \mu &=\pi\cdot(N-1),\ \sigma = 2\sqrt{3} \frac{N-1}{\sqrt{N^2+2}},
\end{split}
\end{equation}
with the width of the peak $\sigma = 2\sqrt{3} \frac{N-1}{\sqrt{N^2+2}}$. This function quickly converges to the constant value $\sigma \xrightarrow{N\rightarrow \infty} 2\sqrt{3}$. When looking at how much space the central peak, containing the solution, occupies relative the single parameter interval we see that it scales as $\frac{\sigma}{2\pi \cdot(N-1)} \sim \frac{1}{N}$. 

The optimization landscape of Target-Kernel Alignment is clearly non-convex, with many local extrema. It is to be expected for the function obtained from composition of cosines. The landscape resembles an interference patter with constructive interference around the solution. Nevertheless, magnitude of local, non-central, maxima decreases with growing $N$ fast enough that the area under the $\mathcal{T}(\mathcal{K})$ function is quickly dominated by the central peak. The flattening of the global optimization landscape with the decreasing width of the solution extremum is typical in quantum machine learning. There are numerous results on the existence of Barren Plateaus {\cite{mcclean2018barren}} which lead exponentially fast to the similar untrainable landscape with increasing number of qubits. The domination of the central peak in the area under the $\mathcal{T}(\mathcal{K})$ is numerically confirmed by comparing the integral of single-peak, Gaussian approximation $\mathcal{G}(x,\mu,\sigma)$ (Equation~\ref{eq:gaussian}) with the integral of $\mathcal{T}(\mathcal{K})$. The integral $\frac{1}{2\pi\cdot(N-1)}\int d\gamma \mathcal{T}(\mathcal{K}) = \langle \mathcal{T}(\mathcal{K}) \rangle$ can be identified with the expected value of the Kernel-Target Alignment.
In Fig. \ref{fig:fit} we show an average value of TA as a function of number of data points $N$. The fitted function confirms the $\sim 1/N$ scaling of the $\langle \mathcal{T}(\mathcal{K})\rangle$ obtained in gaussian approximation. The fitted exponent is close to the expected $-1$ value, the small deviation from it originates from small $N$ values where asymptotic $1/N$ behaviour is not expected.

\section{Numerical study of Target-Kernel alignment expectation value}
It can be argued that increasing the number of data points $N$ in the model is changing the model itself and the effect of the area under the $\mathcal{T}(\mathcal{K})$ comes purely from changing the model while changing $N$.
Therefore, in the numerical experiment we fix the number of available for training data points $N$ but the points are introduced gradually.
In order to always calculate the Kernel-Target Alignment for balanced data, in every iteration of the experiment we add to the training data set two randomly selected points from each class. We start with 2 data points, draw at each iteration subsequent data points and continue to analyse the average value of $\mathcal{T}(\mathcal{K})$. 

In Fig. \ref{fig:real_av}, we render an average value of TA over the parameter space as we draw additional training points in the toy model and the real-life dataset of multispectral images. The real-life data is taken from the 38-Cloud dataset \cite{38-cloud-1, 38-cloud-2}, which consists of multispectral satellite images (acquired by the Landsat-8 satellite) capturing four spectral bands (R, G, B, NIR). In order to compare the analysis with the toy model, only the principal component of the sampled data point was taken. The average value of TA decreases as we include more points in the training dataset. 
The behaviour of the average Target Kernel Alignment for the toy model and real-life data is virtually identical. It indicates that the effect of $\sim 1/N$ decrease in $\mathcal{T}(\mathcal{K})$ value is independent of the data set used. 



\section{Conclusions}

The optimization algorithms in quantum machine learning  are swamped with obstacles \cite{anschuetz2022quantum}. Highly expressive circuits have problems with a flat optimization landscape, which renders gradient based methods useless. We showed that also for circuits with low expressibility, the optimization landscape can become troublesome and dependent on the amount of data used. The Kernel-Target Alignment function possess either many minima or a flat landscape around the vanishingly narrow solution in parameter space. Knowing that the Kernel-Target Alignment value is defined on the interval $[0,1/\sqrt{2}]$ the decrease of its average value indicates that the good optimization solution quickly becomes hard to obtain. Therefore, we recommend to adopt a strategy for the TA optimization even in fairly simple circuits. The best known solution is to exploit the symmetries present in the data \cite{glick2021covariant}. However, most real-life data sets do not exhibit obvious symmetries, hence it is important to look for other approaches to simplify the TA optimization. One possibility could be to design the optimization phase in such a way that we can utilize pretraining on significantly smaller training sets. On the other hand, selecting the appropriate reduced training sets which are highly representative in term of the data structure and its underlying characteristics has been widely researched for support vector machines, and could be effectively deployed in quantum machine learning as well~\cite{nalepa2019selecting}---this constitutes our current research efforts.

\bibliographystyle{IEEEbib}
\bibliography{ref_all}

\subsubsection*{Acknowledgements} 

This work was funded by the European Space Agency,
and supported by the ESA $\Phi$-lab (\url{https://philab.phi.esa.int/}) AI-enhanced 
Quantum Computing for Earth Observation (QC4EO) 
initiative, under ESA contract No. 4000137725/22/NL/GLC/my. 
AM, JM, GC, and FS were supported by the Priority Research 
Areas Anthropocene and Digiworld under the program 
Excellence Initiative – Research University at the 
Jagiellonian University in Krak\'ow. JN was supported 
by the Silesian University of Technology grant for 
maintaining and developing research potential.

\end{document}